\documentclass[10pt, letterpaper]{article}

\usepackage{cogsci}
\cogscifinalcopy

\title{Towards a path dependent account of category fluency}
 
\author{
    {\large \bf David Heineman, Reba Koenen, \and Sashank Varma} \\
    \texttt{\small david.heineman@gatech.edu, rkoenen3@gatech.edu, varma@gatech.edu} \vspace{0.5pt} \\
    School of Interactive Computing, Georgia Tech \\
    Atlanta, GA 30308 USA
}

\begin{document}

\maketitle

\begin{abstract}
Category fluency is a widely studied cognitive task. Two major competing accounts have been proposed as the underlying retrieval mechanism: an optimal foraging process deliberately searching through memory \citep{hills2012optimal} and a random walk sampling from a semantic network \citep{abbott2015random}. Evidence for both accounts has centered around predicting human patch switches, where both existing models of category fluency produce paradoxically identical results. We begin by peeling back the assumptions made by existing models, namely that each named exemplar \textit{only} depends on the previous exemplar, by (i) adding an additional bias to model the category transition probability directly and (ii) relying on a large language model to predict based on the full prior exemplar sequence. Then, we present evidence towards resolving the disagreement between different models of foraging by reformulating them as \textit{sequence generators}. For evaluation, we compare generated category fluency runs to a bank of human-written sequences by utilizing a metric based on $n$-gram overlap. We find that category switch predictors do not necessarily produce human-like sequences; rather, the additional biases used by the \citet{hills2012optimal} model are required to improve generation quality, which is further improved by our category modification. Even generating exclusively with an LLM requires an additional global cue to trigger the patch switching behavior during production. Further tests on only the search process on top of the semantic network highlight the importance of deterministic search in replicating human behavior.  

\textbf{Keywords:} 
category fluency, optimal foraging, large language models, typicality, semantic networks
\end{abstract}

\section{Introduction}
Category fluency tests an individuals' semantic organization and retrieval ability by producing exemplars belonging to a particular category (e.g., `Animals') within a given time limit \citep{gruenewald1980free}. This task is used both to understand cognitive search \citep{troyer1998clustering} and cognitive impairments in executive function such as dementia, schizophrenia or frontal lobe lesions \citep{prudhommeaux-etal-2017-vector}. 
Using human category fluency tests, \citet{hills2012optimal} observed that participants tend to cluster semantically similar exemplars into \textit{patches}, before producing a pause and then a \textit{switch} to a new patch, with a switch indicated by the intermediate time taken to produce an exemplar exceeding the average time to produce all exemplars. 
This patch switching threshold for human response time follows the marginal value theorem (MVT), which states the decision to leave patches occurs when the marginal utility of the current patch falls below the long-term average utility among all patches. MVT is a core principle of optimal foraging in ecology \citep{pyke1977optimal} and information search \citep{pirolli1999information} .

From this observation, two dominant accounts of memory search in category fluency have emerged. The first models retrieval as optimal foraging within semantic space \citep{hills2012optimal}, and which assumes a dual-process search which uses a \textit{local cue} to search within a patch and a \textit{global cue} to trigger patch switches.
The second models retrieval as a random walk through a highly structured space, where the construction of the underlying semantic network leads to the observed foraging behavior \citep{abbott2015random}. Interestingly, both models, constructed with different implicit assumptions about retrieval and the internal human semantic network, exhibit the same MVT behavior when predicting human category switching. 
The current study resolves these competing accounts by relaxing the assumptions of existing models, creating new methods to measure fit to human behavior, and isolating the role of noise in the search process.

We begin by addressing limitations of existing models.
Specifically, both models represent category fluency as a finite state Markov process, where each exemplar only depends on the previous exemplar. 
However, cognitive science and neuroscience data suggests a \emph{hierarchical process} where exemplars are organized by hidden states representing subcategories (e.g., `Pets') \citep{troyer1998clustering}.
We show these subcategories can be exploited to add information into the exemplar production process.
Additionally, we model the task directly with Large Language Models (LLMs), which can directly predict the next exemplar given a full prior sequence of human-written exemplars.
These new formulations allow testing of a non-Markovian account of semantic retrieval, which our experiments reveal better fits the human data.

Next, we show category switch prediction, the primary evidence of category fluency models, produces the same results on all existing and proposed models, showing the difficulty in using switch prediction to differentiate between them.
We improve this evaluation by reframing category fluency \textit{as generation}.
We evaluate generated fluency runs using the 141 human sequences from \citet{hills2012optimal} as a gold-standard dataset. We adapt overlap-based sequence evaluation from machine translation \citep{papineni-etal-2002-bleu} to compare a candidate generation set against the human run distribution. 

Finally, the current study critically re-examines existing accounts of category fluency by separating semantic network construction from the search process. 
We run existing models using both a random and deterministic sampler, showing that a greedy search outperforms a diverse random sampler in producing human-like category fluency sequences.

Our results show that incorporating subcategories into existing models improves the fit to human-like category fluency. We find LLMs are robust category switch predictors yet provide quite poor generation ability. However, LLMs are dramatically improved by adding a global cue to up-weight exemplar frequency, similar to that used in \citet{hills2012optimal}. Further experiments provide evidence that the \citet{hills2012optimal} dual process account of category fluency produces better generations across both existing and proposed models. Our findings further support the theory of semantic retrieval as foraging through memory. Finally, our work provides new modeling, and evaluation tools for category fluency\footnote{We make our data and code publicly available at \url{https://github.com/davidheineman/category-fluency}}.

\subsection{Related Work}
\citet{hills2012optimal} first argued for a foraging process in category fluency, showing that subcategory switches in participant responses corresponded to the marginal value theorem. Specifically, a switch occurred when the intermediate response time (IRT) to produce the next exemplar exceeded the overall mean IRT. This switching behavior was fit by a dual process model defined over a network of pairwise exemplar (semantic) similarities, using a global cue to monitor category switches and a local cue to search within a category. 

Rather than construct the semantic network using embedding similarity, \citet{abbott2015random} observed that a network of free association norms can produce the optimal foraging behavior with a \textit{random walk}: by randomly sampling the next exemplar at each step. In fact, an even stronger fit can be obtained by using a semantic network derived directly from category fluency responses \citep{zemla2018estimating}, suggesting the foraging behavior can be understood as driven by network construction rather than a search process during production. However, it is unclear whether the global cue is indirectly encoded in free association norms, which \citet{jones2015hidden} argues may lead to the parallel (i.e., global and local cue) behavior. The distinction between both accounts of memory organization lies in the assumptions of the semantic network \citep{avery2018comparing}, underscoring the need for new analytical tools for separating the memory and search processes, and for a direct evaluation of models against full runs of human category production. 

Recent work shows that LLMs encode human biases across a variety of cognitive tests, such as content effects when judging syllogisms or performing the Wason selection task, on which LLMs demonstrate human-like failure modes \citep{dasgupta2022language}. Understanding the emergent category structure in language models is also an active area of study \citep{sajjad-etal-2022-analyzing}, but remains challenging as categorization behavior is highly sensitive to the task and prompt construction \citep{suresh2023behavioral}. Here, we present more optimistic results for human category performance, showing the ability of LLMs to match human performance by adding the same global cue used to guide search over the semantic network.
This follows work predicting human biases using LLMs on reading comprehension \citep{wilcox2020predictive} and temporal language processing \citep{goldstein2023temporal}. The current study is the first to analyze LLM behavior using the category fluency task.

\section{Method}
\subsection{Data}

We use the \citet{hills2012optimal} data collected from 141 undergraduates who were given the `animals' superordinate category and asked to list as many exemplars as possible within three minutes. The dataset consists of 5,079 exemplars (369 unique animals). We use the updated category norms of \citet{koenen2022typicality}, which fixed errors in the original \citet{troyer1997clustering} scheme and restructured the 22 subcategories into 16 subcategories to more evenly distribute exemplars.

\subsection{Models of Category Fluency}

In this section, we describe the two dominant models of category fluency, a variation to directly model subcategory switching, and a formulation of LLMs as models of category fluency. The models rely on two mechanisms to guide production: a \textit{local cue} to suggest the next exemplar within a category, and a \textit{global cue} to monitor retrieval and trigger a switch between clusters. The semantic network is formulated as a directed graph, with different search methods as specifying traversal operations across the graph. 
The initial \citet{hills2012optimal} study constructed the network using the pairwise BEAGLE \citep{jones2007representing} similarity between exemplars. By contrast, \citet{abbott2015random} use an unweighted graph of association norms from \citet{nelson2004university}. In our experiments, we replaced BEAGLE similarity with cosine similarity between more recent GloVe \citep{pennington-etal-2014-glove} embeddings, and replicate the same graph of association norms.
Because the network is noisy for less related examples, all edges below a fitted threshold $\epsilon$ are pruned. For our experiments, we follow the comparison study of \citet{avery2018comparing} and use $\epsilon=0.05$. We construct the global cue as an `Animal' node weighted by exemplar frequency. We initially calculated exemplar frequency using the occurrence counts on the Wikipedia corpus reported by \citet{hills2012optimal}. However, we found these to be poorly representative of the animal categorization task (e.g., `human' was the overwhelmingly most typical example, despite being produced in only 9\% of human sequences) and found better performance by replacing the weights by exemplar frequency of occurrence across all 141 participant runs.

\para{Memory Search as Optimal Foraging \citep{hills2012optimal}.} The cue switching model proposes a local cue to estimate the transition probability from the current exemplar to the next in the semantic network $P(x_{n+1}|x_n)$ corresponding to the cosine similarity between the GloVe embeddings of each exemplar, and a global cue to estimate the probability of the next exemplar $P(x_{n+1}|`animal')$ estimated using the exemplar frequency across the \citet{hills2012optimal} participant runs.
Formally, the network is fit to a local and global cue parameter $\beta_l$ and $\beta_g$ respectively:

\vspace{-10pt}
$$P(x_{n+1} | x_n) = \underbrace{P(x_{n+1}|x_{n})^{\beta_l}}_{\text{local cue}} \times \underbrace{P(x_{n+1}|\text{`animal'})^{\beta_g}}_{\text{global cue}}$$

\noindent\textbf{Memory Search as a Random Walk \citep{abbott2015random}.} The random walk approach is a reduction of the above cue switching model, specifically relying entirely on the network (i.e., setting $\beta_g=0$), and representing the local transition probability by uniformly selecting any outgoing connections from $x_n$, $P(x_{n+1}|x_n)=\mathds{1}[x_{n}\rightarrow x_{n+1}]$. As there is no scalar cue probability, identifying category switches requires sampling full sequences in a random walk and recording the number of intermediary nodes until the node is revisited. Formally, let $e(i)$ represent the total path length on the $i$-th visit to exemplar $e$, therefore $\text{IRT}(e)=e(1)-e(0)$, or the length of the first cycle containing $e$ on the path. \citet{abbott2015random} experimented with a semantic network constructed from word association data \citep{nelson2004university}, relying on the complexity of the semantic network to produce foraging behavior. We test networks that replicate using these association data and that are novel in using GloVe pairwise similarity.

\begin{figure}[t!]
\centering
\includegraphics[width=0.5\textwidth]{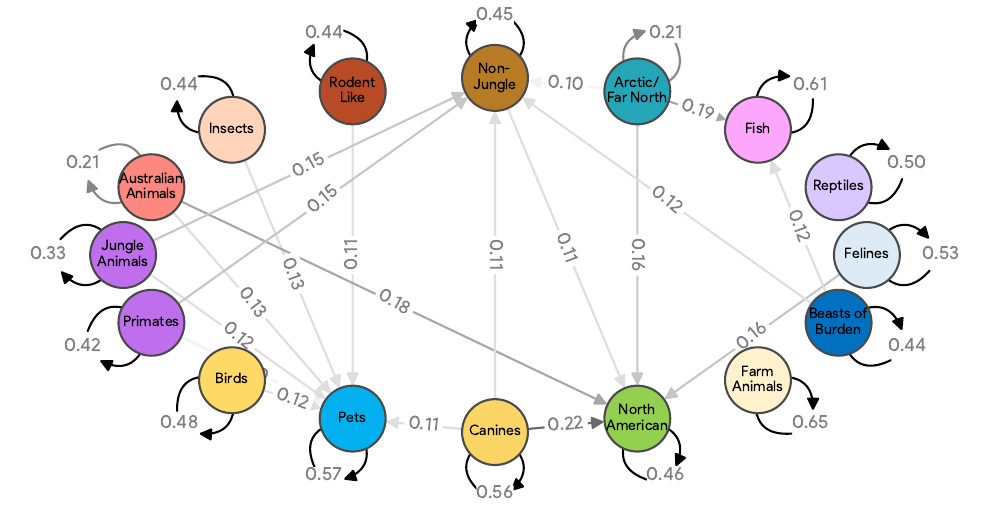}
\setlength{\belowcaptionskip}{-7pt}
\setlength{\abovecaptionskip}{-5pt}
\captionof{figure}{Category transition strengths $P(c_{n+1}|c_n) > 0.1$ calculated using human switches from \protect\citet{hills2012optimal} and the category schema proposed in \protect\citet{koenen2022typicality}.}
\label{fig:category_transitions}
\end{figure}

\para{Memory Search Incorporating the Subcategory Cue.}
Because humans are far more likely to jump between semantically similar patches \citep{troyer1997clustering}, we explore directly incorporating subcategory information into the category fluency model. In this more hierarchical model, the local cue searches exemplars within a patch, the global cue estimates exemplar typicality for the whole task, and the new subcategory cue models \textit{relationships between clusters}. Thus, we augment the cue switching model with a subcategory cue, defined as the frequency of the category of the next exemplar given the category of the current exemplar:

\vspace{-5pt}
{\small
$$P(x_{n+1} | x_n) =  \underbrace{P(x_{n+1}|x_n)^{\beta_l}}_{\text{local cue}} \times \underbrace{P(x_{n+1}|\text{`animal'}) ^{\beta_g}}_{\text{global cue}}\times \underbrace{P(c_{n+1}|c_n) ^{\beta_c}}_{\text{subcategory cue}}$$
}

\noindent Figure \ref{fig:category_transitions} illustrates how learned transition probabilities between exemplar categories add inertia to traversing out of a category run and encourage cue models to jump to semantically similar subordinate patches. While this relationship may be implicitly encoded in the GloVe similarity, the subcategory cue explicitly accounts for hidden subcategory states.

\begin{figure}[t!]
\centering
\includegraphics[width=0.45\textwidth]{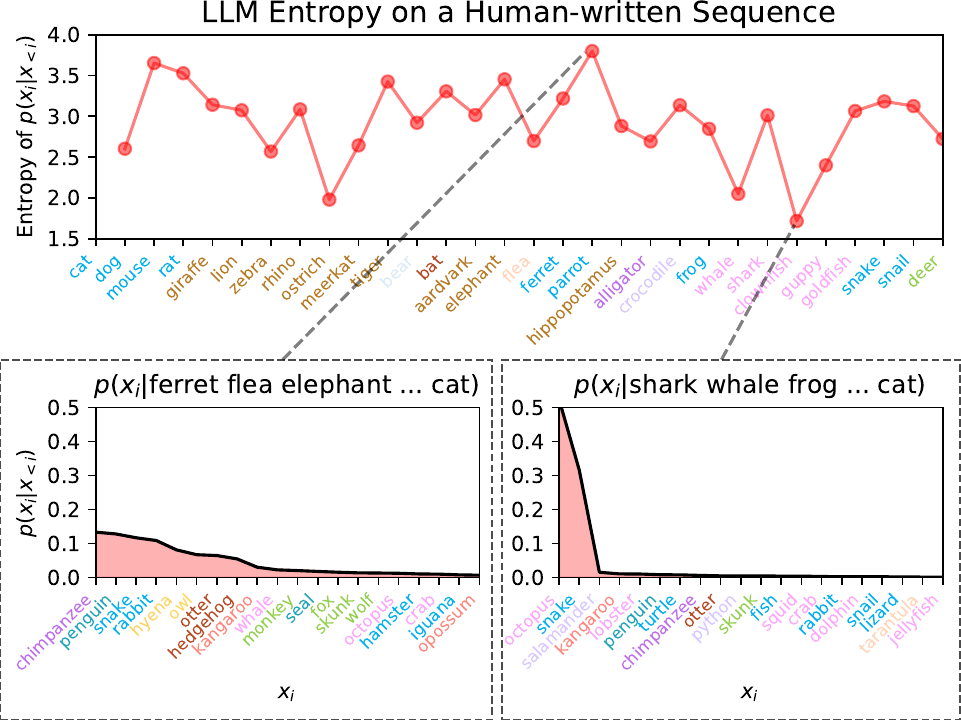}
\setlength{\belowcaptionskip}{-8pt}
\captionof{figure}{LLM entropy over a human-written sequence from \protect\citet{hills2012optimal}, with each example colored by its categorization coding. We observe high entropy in the next-token prediction (bottom left) corresponds to a jump between semantic patches, and a low entropy (bottom right) corresponds to exploration within a patch.}
\label{fig:llm_entropy}
\end{figure}

\para{Memory Search on a Fully Conditioned Sequence}
As the above models operate on top of a defined semantic space (i.e., BEAGLE, GloVe), they are only conditioned on the current exemplar $x_n$, requiring assumptions about memory search procedures. Therefore, a goal of the current research is to evaluate whether LLMs replicate human category fluency behavior without additional modifications. 
Autoregressive LLMs are trained to predict a distribution across tokens given by $P(x_{n+1}|x_{n}, \dots, x_0)$. We aim to directly apply this distribution as a stand-in for category fluency by providing human-written prefixes and decoding the next exemplar distribution using an LLM. For our experiments we use the open-source Llama 2 Chat 7B model \citep{touvron2023llama}, fine-tuned to follow natural language instructions.
We chose Llama 2 because we have full access to the prediction probabilities, its widespread use in NLP, and the lack of equivalent models at the time of writing.
When performing experiments, we prepend a description of the category fluency task, decode with an early-stopping beam search \citep{huang-etal-2017-finish} to recover full exemplars along with a probability distribution, and remove any generations outside the \citet{koenen2022typicality} categorization scheme. Figure \ref{fig:llm_entropy} shows the Llama 2 Chat LLM entropy across a human-written sequence, where we later show that averaged across all category switches, low confidence implies the model has exhausted an existing patch. In later experiments, we use the LLM prediction as the local cue in the original \citet{hills2012optimal} model, as we find that this improves the prediction ability.

\subsection{Evaluation on Category Fluency Sequences}

Evaluation of category fluency has focused on the category switching operation: attempting to model the mean value theorem, or the spike in human response time just before switching to a new category. However, a new approach that is now available is to model the category fluency task directly, testing the generative power of LLMs. To do this, we begin a walk by starting at the `animal' node in the semantic network and sample from the next exemplar probability distribution given by each node. Once a set of candidate sequences is generated, we measure their fit to a bank of human sequences by adapting BLEU, an overlap-based metric developed for evaluating machine translation \citep{papineni-etal-2002-bleu}. BLEU measures $n$-gram overlap to a set of references:

\vspace{-10pt}
{\small
$$\text{BLEU} = \underbrace{\Big(\prod_{i=1}^{3} \text{precision}_i\Big)^{1/3}}_{i\text{-gram overlap for } i\,\in\,\{1,2,3\}}\times \underbrace{\vphantom{\prod_i^3}\min\Big(1,\exp\big(1-\frac{\text{{\tiny human length}}}{\text{{\tiny generation length}}}\big)\Big)}_{\text{brevity penalty}}$$
}
\vspace{-5pt}

\noindent where $\text{precision}_i$ is the overlap of unigram/bigram/trigrams among all human references, formally defined as 
$\text{precision}_i = {\sum_{h\in\mathcal{H}}\sum_{i\in h}\min(m^i_{h}, m^i_{gen})}/{ \sum_{h\in\mathcal{H}}\sum_{i\in h} m^{i}_{h}}$
where $\mathcal{H}$ represents the set of all experimental human sequences and $m_i$ represents the number of $i$-grams in either the human sequence $m_h$ or generation $m_{gen}$, with an additional brevity penalty to prevent arbitrarily long sequences containing all possible $i$-grams. 
We report two measures of BLEU: \textit{Exemplar BLEU}, the overlap of the full generated sequence to human references, and \textit{Category BLEU}, the overlap of the categories of generated exemplars to the human category coding collected by \citet{koenen2022typicality}.

\subsection{Category Fluency Models as Sequence Generators}

Because we are drawing samples from the exemplar probability distribution, the choice of search technique has a high impact on the sequence quality \citep{avery2018comparing} and can be used to isolate the interaction between search algorithms and the semantic network. A deterministic search attempts to maximize the cue probability, such as using greedy search to select the most likely exemplar at each step or an exhaustive search to find the most likely joint probability between exemplars. As finding the sequence of maximum likelihood would require testing each permutation of exemplars, we instead use beam search to estimate high probability paths by exploring multiple candidate sequences simultaneously, pruning the search paths of low probability. 

\para{Random Walks in Semantic Space.} \citet{abbott2015random} argues sampling behavior involves performing random walks over a patchy environment. Because sampling from a distribution introduces an ability to control the entropy of the next exemplar prediction with a temperature parameter $\tau$ \citep{ackley1985learning}, we adjust the impact of noise on the cue switching model, with a random walk being equivalent to sampling from a uniform distribution ($\tau=\infty$) of only a local cue. 
This allows progressively testing whether the explore / exploit behavior in semantic foraging is a product of a deliberate prediction of the model (deterministic search) or a byproduct of random exploration over a network (noisier search).

\section{Results}
\begin{figure}[t!]
\centering
\includegraphics[width=0.3\textwidth]{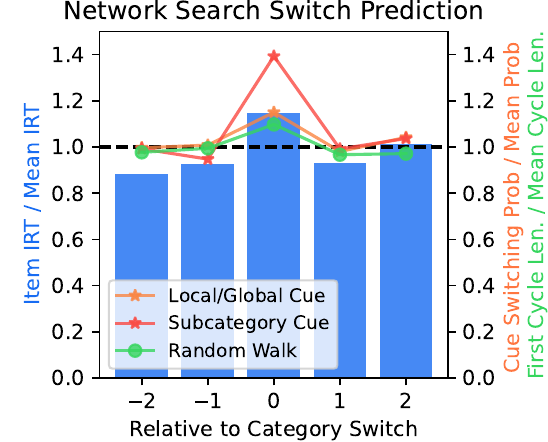}
\captionof{figure}{Response time on \protect\citet{hills2012optimal}, along with network search switch prediction models. We find the subcategory cue exaggerates the Hills switch prediction model.}
\label{fig:subcategory_switch}
\end{figure}
\begin{figure}[t!]
\centering
\includegraphics[width=0.3\textwidth]{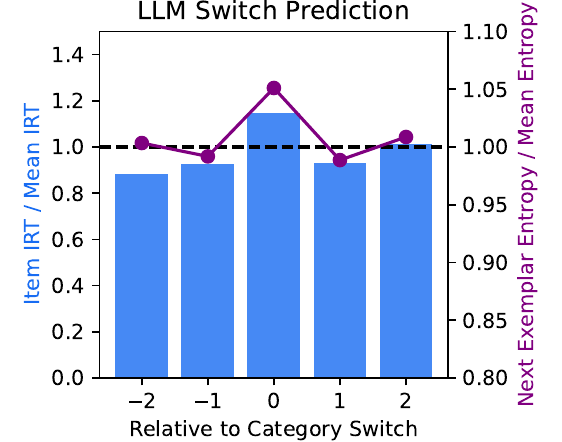}
\setlength{\belowcaptionskip}{-7pt}
\captionof{figure}{Category switch prediction using Llama 2 Chat entropy across exemplars given human category fluency runs.}
\label{fig:llm_switch}
\end{figure}

We begin by revisiting the category switch prediction experiments primarily used as evidence for the \citet{hills2012optimal} and \citet{abbott2015random} models. Then, we evaluate the quality of sequence generation using each model. We conclude by comparing deterministic and stochastic search approaches to semantic foraging.

\subsection{Category Switch Prediction}

We report the intermediate response time (IRT) for the exemplars immediately before and immediately after a category switch in Figure \ref{fig:subcategory_switch}; these are the behavioral data, and they show clear evidence of MVT behavior. The red line shows the total probability of switching to an out-of-category exemplar over the mean probability for the full run for the \citet{hills2012optimal} switching model; the orange line shows this measure for our proposed subcategory variant. Both models effectively predict a category switch, with a spike in probability when producing the first exemplar in the next category. Our subcategory model exaggerates the relationship already learned by the cue model. This demonstrates that adding this additional hierarchical information enables the model to better capture the same MVT switching relationship. 

Figure \ref{fig:subcategory_switch} compares the probability of only the next single human exemplar. However, the next human exemplar occurs in only 30\% of LLM candidate generations. Therefore, the entropy of the distribution of LLM candidate generations is a richer indication of confidence than the exact candidate. Figure \ref{fig:llm_switch} shows the results from our LLM setup. The purple line measures the entropy of the next exemplar distribution given the human sequences. We observe the same MVT behavior, where a jump in the entropy corresponds to a patch switch in the human data. 
This finding shows that an unmodified LLM entropy also encodes the MVT observation to predict human patch switching. It also demonstrates three different accounts of category fluency produce identical findings. The next sections rely on stricter constraints to distinguish the models.

\subsection{Category Fluency Generation}

Isolating model design choices, and more generally choosing between models based on switch prediction performance alone, is challenging. We therefore perform a more holistic evaluation of semantic networks’ generation ability by sampling 1000 sequences and averaging the BLEU scores given all human category runs as human references; see Table \ref{table:generation_eval} for the results, where a higher BLEU score indicates a better overlap with human category fluency data. 
We report `Exemplar BLEU' for the generated exemplars and `Category BLEU' for the categories of generated exemplars to the categories of human sequences. A higher Exemplar BLEU indicates better fit to human exemplar sequences while Category BLEU indicates the correct ordering of category visits. 
Both the out-of-the-box random walk and cue switching models are ineffective sequence generators. However, when the Wikipedia word frequencies are replaced with exemplar occurrences in the \citet{hills2012optimal} data (`Gold Freq.') to create a more accurate global cue estimation, the cue switching models performed substantially better. 

\begin{table}[t!]
\centering
\small
\begin{tabular}{L{76pt}C{33pt}C{33pt}C{23pt}C{23pt}}
\toprule

\multirow{2}{*}{\textit{Semantic Network}} & 
\makebox[0pt][c]{\parbox[t]{33pt}{\centering\textbf{Exemplar\newline BLEU}}} & 
\makebox[0pt][c]{\parbox[t]{33pt}{\centering\textbf{Category\newline BLEU}}} & 
\makebox[0pt][c]{\parbox[t]{36pt}{\centering\textbf{Avg. Run \newline Length}}} & 
\textbf{\% Switch} \\

\midrule
\multicolumn{5}{l}{\textit{Free Association Norms}} \\
Random Walk & 0.079 & 0.801 & 2.16 & 44.8 \\
\midrule
\multicolumn{5}{l}{\textit{Pairwise GloVe Similarity}} \\
Random Walk & 0.075 & 0.658 & 2.37 & 40.8 \\
Local + Global Cue & 0.062 & 0.569 & 2.93 & 32.8 \\
L + G (Gold Freq.) & 0.122 & 0.783 & 1.95 & 49.8 \\
L + G + Subcategory & \textit{0.182} & \textit{0.851} & 4.54 & 21.4 \\
\midrule
\multicolumn{5}{l}{\textit{Network-free}} \\
LLM & 0.036 & 0.469 & 1.23 & 65.3 \\
LLM + Global Cue & \textbf{0.214} & \textbf{0.967} & 2.53 & 39.3 \\
Human Sequences & 0.220 & 0.922 & 2.14 & 44.6 \\
\bottomrule
\end{tabular}
\setlength{\belowcaptionskip}{-10pt}
\caption{Results comparing sequence generations to human experiments across category fluency models. The addition of a subcategory cue to the \protect\citet{hills2012optimal} model and the global cue to the LLM improves their fit to human sequences.}
\label{table:generation_eval}
\end{table}
\begin{figure}[t!]
\centering
\includegraphics[width=0.4\textwidth]{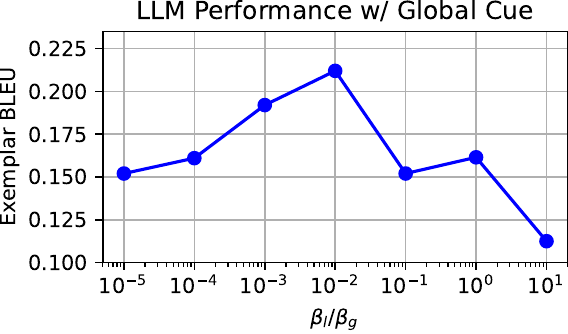}
\setlength{\abovecaptionskip}{4pt}
\setlength{\belowcaptionskip}{-7pt}
\captionof{figure}{Llama 2 7B Chat performance across global cue strengths. Balancing the strength of the local and global cue is necessary for LLMs to produce human-like category runs.}
\label{fig:llm_cue_strength}
\end{figure}

Interestingly, the subcategory switching model further improves fit to both the exemplar and category BLEU scores (i.e., orderings). However, the average category run length (i.e., number of exemplars) and the percentage of switches both begin to diverge from human performance.

Finally, we evaluate performance for LLM-generated category runs. We find that allowing the language model to freely generate leads to poorer performance than any walking model on a semantic network. However, once a global cue is added to the language model predictions by up-weighting exemplars using the exemplar occurrences from the \citet{hills2012optimal} human data (`+ Global Cue'), the produced generations are substantially higher quality as measured by exemplar and category  BLEU performance. 
This indicates the efficiency of stricter assumptions over the semantic network (subcategory cue) and the necessity of a cue-like model both for exploration over a semantic network and the LLM, which is directly applied to the task.
We further evaluate the impact of the global cue in Figure \ref{fig:llm_cue_strength}, where we test different ratios of global-/local-cue strength which demonstrates both the sensitivity and necessity of the two mechanisms in LLM generation. 
A model without a local cue ($\beta_l/\beta_g\approx 0$) is unable to produce human-like sequences on its own, while a model with too strong of a global cue ($\beta_l/\beta_g> 5$) is also not able to balance the local exploitation, global exploration relationship required for human-like category fluency.
This gives evidence for the cue-based model as providing executive control for LLM predictions, showing the \citet{hills2012optimal} account of memory foraging is important for replicating human-like category fluency with LLMs.

\begin{figure*}[t!]
\centering
\includegraphics[width=0.86\textwidth]{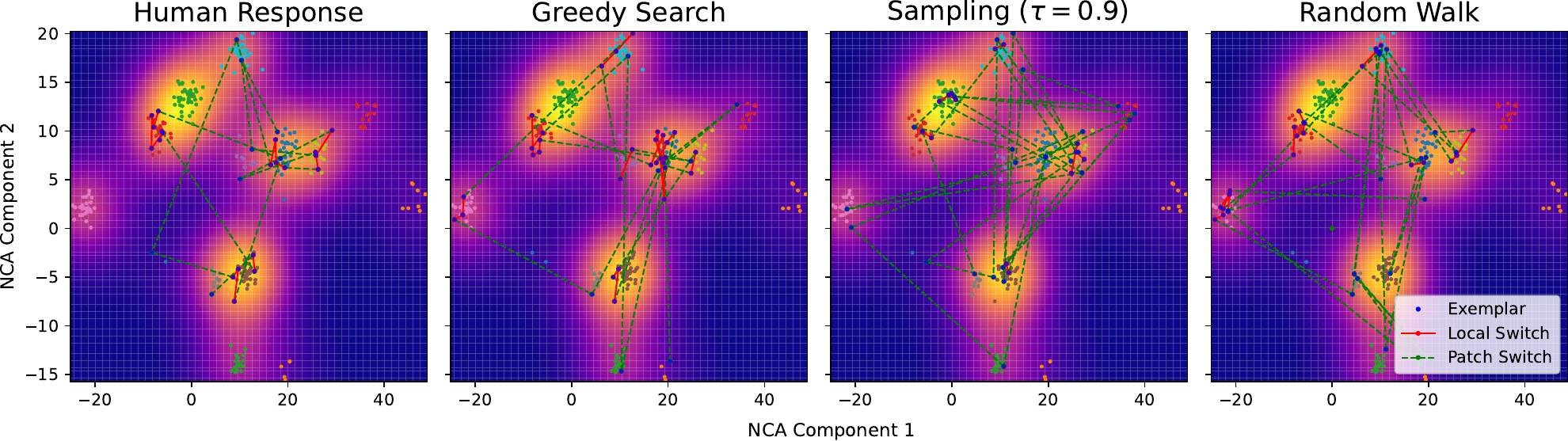}
\setlength{\belowcaptionskip}{-8pt}
\setlength{\abovecaptionskip}{5pt}
\caption{Search paths using the \protect\citet{hills2012optimal} cue switching model on a density map of exemplars projected from high-dimensional GloVe embedding space using neighborhood component analysis \protect\citep{goldberger2004neighbourhood}, colored by category norms from \protect\citet{koenen2022typicality}. The local retrieval and global patch switching process is lost in noisier search methods.}
\label{fig:search_paths}
\end{figure*}

\subsection{Comparing Search Methods as Semantic Foragers}

Models of category fluency disagree on the degree to which human memory foraging is performed during production versus as a side effect of semantic network construction. In this section, we isolate the search process from the semantic network to test the extent to which stochastic noise contributes to category switching behavior. To illustrate this difference in search methods, we generate a single category fluency run with different search methods and plot out the exemplars in Figure \ref{fig:search_paths} by projecting the exemplars onto the GloVe semantic space. We highlight transitions between exemplars in the same category (red) and transitions between exemplars in different categories (green). The human response (left) appears to alternate between search within patches or similar patches, and a few jumps between patches. The greedy search appears to replicate this behavior the closest, with sampling ($\tau=0.9$) and random walk ($\tau=\infty$) performing far too many jumps. However, this large difference in behavior solely due to the search process indicates a need to test the generative ability of the search process in isolation to understand whether stochastic search methods, which are robust switch predictors (Figure \ref{fig:subcategory_switch}), imply a fit to sequence generation.

We therefore test search techniques on the \citet{hills2012optimal} cue model using the \textit{same} GloVe semantic network, with noise injected during the sampling process. Table \ref{table:search_eval} reports the results. 
Our evaluation replicates the previous section, where 1000 generations are sampled from each model and evaluated against the 141 human sequences to calculate the final Exemplar BLEU score. 
We observe deterministic search techniques (i.e., greedy, beam search) report higher sequence scores across all model constructions. While the artificial noise in stochastic generations should benefit by allowing more overlap opportunities to fit the natural variance in human response, any amount of noise degrades quality in practice, even at colder temperatures (e.g., $\tau=0.1$). As the temperature increases, meaning the probability distribution becomes closer to random sampling performed in \citet{abbott2015random}, the Exemplar BLEU performance approaches the same random walk performance shown in the free association norms in Table \ref{table:generation_eval}. Additionally, Exemplar BLEU scores for each search technique are consistent among different model cues choices 
(with the exception of beam search, which created high likelihood, but degenerate, sequences), further highlighting that while a random sampler may predict the switching behavior, deterministic search techniques are crucial for producing human-like runs.

\section{Discussion}
We revisit existing accounts of category fluency, proposing new models and evaluation techniques to isolate the network and search mechanisms involved in semantic foraging. Our first result replicated the category switch prediction, already reported by \citet{hills2012optimal} and \citet{abbott2015random}, with our new models of category fluency, showing the value of directly modeling subcategory transition probabilities, and that the LLM entropy measure also shows the same thresholded function.
In all cases, the additional subcategory cue constraint successfully incorporated the switching behavior as an inductive bias in our model, leading to a better fit to switch prediction. Our category switching results also serves as a critique of switch prediction as evidence of any particular account of category fluency; in fact, multiple accounts produce the same result. This under-determinism motivated a test of category fluency models for generating full human sequences. This new formulation of evaluation as sequence generation allowed independently testing the network between exemplars and the search method used to draw sequences from the network, motivating the two latter experiments.

\begin{table}[t]
\centering
\small

\begin{tabular}{L{2.5cm}C{1.3cm}C{1.3cm}C{1.7cm}}
\toprule
\multirow{2}{*}{\textit{Search Method}} & \textbf{Local Cue Only} & \textbf{Local + Global} & \textbf{L + G + Subcategory} \\
\midrule

Greedy & \textbf{0.18} & \textbf{0.21} & \textbf{0.23} \\
Beam Search & 0.00 & 0.00 & 0.14 \\
Sampling ($\tau=0.1$) & 0.12 & 0.21 & 0.23 \\
Sampling ($\tau=0.9$) & 0.06 & 0.06 & 0.06 \\
Sampling ($\tau=1.5$) & 0.05 & 0.05 & 0.05 \\
Random Walk & 0.08 & 0.08 & 0.08 \\

\bottomrule
\end{tabular}

\setlength{\belowcaptionskip}{-15pt}
\caption{Exemplar BLEU for category fluency models using semantic search methods. Deterministic search methods rely on the fluency model to express foraging behavior.}
\label{table:search_eval}
\end{table}

Our second result tests the generative power of networks and reports (i) stricter modeling assumptions, specifically our subcategory cue model, improve the fit to the ordering of category exploration, and when to trigger a switching decision; and (ii) our baseline LLM does not produce human-like category sequences, but only works once we fit the same local / global cue used in the original \citet{hills2012optimal} model; in fact its generations are more powerful then the network search-based techniques. Framing category fluency as generation thus demonstrates that human-like categorizations can be extracted from LLMs with a simple additional global cue.

Our third result re-evaluates the random walk hypothesis by comparing search techniques across a range of temperatures on each network, with a more `heated' walk shifting generation closer to a random walk. Our results provide further evidence that the explore / exploit behavior between patches is a product of a deliberate switching mechanism, rather than noise during the sampling process. Both results are enabled by our adaptation of the overlap-based BLEU score, which allowed us to compare each generation to human references representing the distribution of human responses. When testing semantic networks as generators, we show random walks do not produce category fluency.

\newpage

\bibliographystyle{apacite}
\setlength{\bibleftmargin}{.125in}
\setlength{\bibindent}{-\bibleftmargin}
\bibliography{custom}

\begin{thebibliography}{}

\bibitem [\protect \citeauthoryear {%
Abbott%
, Austerweil%
\BCBL {}\ \BBA {} Griffiths%
}{%
Abbott%
\ \protect \BOthers {.}}{%
{\protect \APACyear {2015}}%
}]{%
abbott2015random}
\APACinsertmetastar {%
abbott2015random}%
\begin{APACrefauthors}%
Abbott, J\BPBI T.%
, Austerweil, J\BPBI L.%
\BCBL {}\ \BBA {} Griffiths, T\BPBI L.%
\end{APACrefauthors}%
\unskip\
\newblock
\APACrefYearMonthDay{2015}{}{}.
\newblock
{\BBOQ}\APACrefatitle {Random walks on semantic networks can resemble optimal foraging.} {Random walks on semantic networks can resemble optimal foraging.}{\BBCQ}
\newblock
\APACjournalVolNumPages{{Psychological Review}}{122}{3}{558--569}.
\PrintBackRefs{\CurrentBib}

\bibitem [\protect \citeauthoryear {%
Ackley%
, Hinton%
\BCBL {}\ \BBA {} Sejnowski%
}{%
Ackley%
\ \protect \BOthers {.}}{%
{\protect \APACyear {1985}}%
}]{%
ackley1985learning}
\APACinsertmetastar {%
ackley1985learning}%
\begin{APACrefauthors}%
Ackley, D\BPBI H.%
, Hinton, G\BPBI E.%
\BCBL {}\ \BBA {} Sejnowski, T\BPBI J.%
\end{APACrefauthors}%
\unskip\
\newblock
\APACrefYearMonthDay{1985}{}{}.
\newblock
{\BBOQ}\APACrefatitle {A learning algorithm for Boltzmann machines} {A learning algorithm for boltzmann machines}.{\BBCQ}
\newblock
\APACjournalVolNumPages{Cognitive science}{9}{1}{147--169}.
\PrintBackRefs{\CurrentBib}

\bibitem [\protect \citeauthoryear {%
Avery%
\ \BBA {} Jones%
}{%
Avery%
\ \BBA {} Jones%
}{%
{\protect \APACyear {2018}}%
}]{%
avery2018comparing}
\APACinsertmetastar {%
avery2018comparing}%
\begin{APACrefauthors}%
Avery, J.%
\BCBT {}\ \BBA {} Jones, M\BPBI N.%
\end{APACrefauthors}%
\unskip\
\newblock
\APACrefYearMonthDay{2018}{}{}.
\newblock
{\BBOQ}\APACrefatitle {Comparing models of semantic fluency: Do humans forage optimally, or walk randomly?} {Comparing models of semantic fluency: Do humans forage optimally, or walk randomly?}{\BBCQ}
\newblock
\BIn{} \APACrefbtitle {{Proceedings of the Annual Meeting of the Cognitive Science Society}.} {{Proceedings of the Annual Meeting of the Cognitive Science Society}.}
\PrintBackRefs{\CurrentBib}

\bibitem [\protect \citeauthoryear {%
Dasgupta%
\ \protect \BOthers {.}}{%
Dasgupta%
\ \protect \BOthers {.}}{%
{\protect \APACyear {2022}}%
}]{%
dasgupta2022language}
\APACinsertmetastar {%
dasgupta2022language}%
\begin{APACrefauthors}%
Dasgupta, I.%
, Lampinen, A\BPBI K.%
, Chan, S\BPBI C.%
, Creswell, A.%
, Kumaran, D.%
, McClelland, J\BPBI L.%
\BCBL {}\ \BBA {} Hill, F.%
\end{APACrefauthors}%
\unskip\
\newblock
\APACrefYearMonthDay{2022}{}{}.
\newblock
{\BBOQ}\APACrefatitle {Language models show human-like content effects on reasoning} {Language models show human-like content effects on reasoning}.{\BBCQ}
\newblock
\APACjournalVolNumPages{arXiv preprint arXiv:2207.07051}{}{}{}.
\PrintBackRefs{\CurrentBib}

\bibitem [\protect \citeauthoryear {%
Goldberger%
, Hinton%
, Roweis%
\BCBL {}\ \BBA {} Salakhutdinov%
}{%
Goldberger%
\ \protect \BOthers {.}}{%
{\protect \APACyear {2004}}%
}]{%
goldberger2004neighbourhood}
\APACinsertmetastar {%
goldberger2004neighbourhood}%
\begin{APACrefauthors}%
Goldberger, J.%
, Hinton, G\BPBI E.%
, Roweis, S.%
\BCBL {}\ \BBA {} Salakhutdinov, R\BPBI R.%
\end{APACrefauthors}%
\unskip\
\newblock
\APACrefYearMonthDay{2004}{}{}.
\newblock
{\BBOQ}\APACrefatitle {Neighbourhood components analysis} {Neighbourhood components analysis}.{\BBCQ}
\newblock
\APACjournalVolNumPages{{Advances in Neural Information Processing Systems}}{17}{}{}.
\PrintBackRefs{\CurrentBib}

\bibitem [\protect \citeauthoryear {%
Goldstein%
\ \protect \BOthers {.}}{%
Goldstein%
\ \protect \BOthers {.}}{%
{\protect \APACyear {2023}}%
}]{%
goldstein2023temporal}
\APACinsertmetastar {%
goldstein2023temporal}%
\begin{APACrefauthors}%
Goldstein, A.%
, Ham, E.%
, Schain, M.%
, Nastase, S.%
, Zada, Z.%
, Dabush, A.%
\BDBL {}Hasson, U.%
\end{APACrefauthors}%
\unskip\
\newblock
\APACrefYearMonthDay{2023}{}{}.
\newblock
{\BBOQ}\APACrefatitle {The Temporal Structure of Language Processing in the Human Brain Corresponds to The Layered Hierarchy of Deep Language Models} {The temporal structure of language processing in the human brain corresponds to the layered hierarchy of deep language models}.{\BBCQ}
\newblock
\APACjournalVolNumPages{arXiv preprint arXiv:2310.07106}{}{}{}.
\PrintBackRefs{\CurrentBib}

\bibitem [\protect \citeauthoryear {%
Gruenewald%
\ \BBA {} Lockhead%
}{%
Gruenewald%
\ \BBA {} Lockhead%
}{%
{\protect \APACyear {1980}}%
}]{%
gruenewald1980free}
\APACinsertmetastar {%
gruenewald1980free}%
\begin{APACrefauthors}%
Gruenewald, P\BPBI J.%
\BCBT {}\ \BBA {} Lockhead, G\BPBI R.%
\end{APACrefauthors}%
\unskip\
\newblock
\APACrefYearMonthDay{1980}{}{}.
\newblock
{\BBOQ}\APACrefatitle {The free recall of category examples.} {The free recall of category examples.}{\BBCQ}
\newblock
\APACjournalVolNumPages{Journal of Experimental Psychology: Human Learning and Memory}{6}{3}{225}.
\PrintBackRefs{\CurrentBib}

\bibitem [\protect \citeauthoryear {%
Hills%
, Jones%
\BCBL {}\ \BBA {} Todd%
}{%
Hills%
\ \protect \BOthers {.}}{%
{\protect \APACyear {2012}}%
}]{%
hills2012optimal}
\APACinsertmetastar {%
hills2012optimal}%
\begin{APACrefauthors}%
Hills, T\BPBI T.%
, Jones, M\BPBI N.%
\BCBL {}\ \BBA {} Todd, P\BPBI M.%
\end{APACrefauthors}%
\unskip\
\newblock
\APACrefYearMonthDay{2012}{}{}.
\newblock
{\BBOQ}\APACrefatitle {Optimal foraging in semantic memory.} {Optimal foraging in semantic memory.}{\BBCQ}
\newblock
\APACjournalVolNumPages{{Psychological Review}}{119}{2}{431}.
\PrintBackRefs{\CurrentBib}

\bibitem [\protect \citeauthoryear {%
Huang%
, Zhao%
\BCBL {}\ \BBA {} Ma%
}{%
Huang%
\ \protect \BOthers {.}}{%
{\protect \APACyear {2017}}%
}]{%
huang-etal-2017-finish}
\APACinsertmetastar {%
huang-etal-2017-finish}%
\begin{APACrefauthors}%
Huang, L.%
, Zhao, K.%
\BCBL {}\ \BBA {} Ma, M.%
\end{APACrefauthors}%
\unskip\
\newblock
\APACrefYearMonthDay{2017}{}{}.
\newblock
{\BBOQ}\APACrefatitle {When to Finish? {O}ptimal Beam Search for Neural Text Generation (modulo beam size)} {When to finish? {O}ptimal beam search for neural text generation (modulo beam size)}.{\BBCQ}
\newblock
\BIn{} \APACrefbtitle {{Proceedings of the Conference on Empirical Methods in Natural Language Processing}} {{Proceedings of the Conference on Empirical Methods in Natural Language Processing}}\ (\BPGS\ 2134--2139).
\PrintBackRefs{\CurrentBib}

\bibitem [\protect \citeauthoryear {%
Jones%
, Hills%
\BCBL {}\ \BBA {} Todd%
}{%
Jones%
\ \protect \BOthers {.}}{%
{\protect \APACyear {2015}}%
}]{%
jones2015hidden}
\APACinsertmetastar {%
jones2015hidden}%
\begin{APACrefauthors}%
Jones, M\BPBI N.%
, Hills, T\BPBI T.%
\BCBL {}\ \BBA {} Todd, P\BPBI M.%
\end{APACrefauthors}%
\unskip\
\newblock
\APACrefYearMonthDay{2015}{}{}.
\newblock
{\BBOQ}\APACrefatitle {Hidden processes in structural representations: A reply to {A}bbott, {A}usterweil, and {G}riffiths (2015).} {Hidden processes in structural representations: A reply to {A}bbott, {A}usterweil, and {G}riffiths (2015).}{\BBCQ}
\newblock
\APACjournalVolNumPages{American Psychological Association}{}{}{}.
\PrintBackRefs{\CurrentBib}

\bibitem [\protect \citeauthoryear {%
Jones%
\ \BBA {} Mewhort%
}{%
Jones%
\ \BBA {} Mewhort%
}{%
{\protect \APACyear {2007}}%
}]{%
jones2007representing}
\APACinsertmetastar {%
jones2007representing}%
\begin{APACrefauthors}%
Jones, M\BPBI N.%
\BCBT {}\ \BBA {} Mewhort, D\BPBI J.%
\end{APACrefauthors}%
\unskip\
\newblock
\APACrefYearMonthDay{2007}{}{}.
\newblock
{\BBOQ}\APACrefatitle {Representing word meaning and order information in a composite holographic lexicon.} {Representing word meaning and order information in a composite holographic lexicon.}{\BBCQ}
\newblock
\APACjournalVolNumPages{{Psychological Review}}{114}{1}{1}.
\PrintBackRefs{\CurrentBib}

\bibitem [\protect \citeauthoryear {%
Koenen%
, Hayden%
, Herman%
\BCBL {}\ \BBA {} Varma%
}{%
Koenen%
\ \protect \BOthers {.}}{%
{\protect \APACyear {2022}}%
}]{%
koenen2022typicality}
\APACinsertmetastar {%
koenen2022typicality}%
\begin{APACrefauthors}%
Koenen, R.%
, Hayden, B\BPBI Y.%
, Herman, A\BPBI B.%
\BCBL {}\ \BBA {} Varma, S.%
\end{APACrefauthors}%
\unskip\
\newblock
\APACrefYearMonthDay{2022}{}{}.
\newblock
{\BBOQ}\APACrefatitle {Typicality gradients in the category fluency task} {Typicality gradients in the category fluency task}.{\BBCQ}
\newblock
\BIn{} \APACrefbtitle {{Proceedings of the Annual Meeting of the Cognitive Science Society}} {{Proceedings of the Annual Meeting of the Cognitive Science Society}}\ (\BVOL~44).
\PrintBackRefs{\CurrentBib}

\bibitem [\protect \citeauthoryear {%
Nelson%
, McEvoy%
\BCBL {}\ \BBA {} Schreiber%
}{%
Nelson%
\ \protect \BOthers {.}}{%
{\protect \APACyear {2004}}%
}]{%
nelson2004university}
\APACinsertmetastar {%
nelson2004university}%
\begin{APACrefauthors}%
Nelson, D\BPBI L.%
, McEvoy, C\BPBI L.%
\BCBL {}\ \BBA {} Schreiber, T\BPBI A.%
\end{APACrefauthors}%
\unskip\
\newblock
\APACrefYearMonthDay{2004}{}{}.
\newblock
{\BBOQ}\APACrefatitle {The University of South Florida free association, rhyme, and word fragment norms} {The university of south florida free association, rhyme, and word fragment norms}.{\BBCQ}
\newblock
\APACjournalVolNumPages{{Behavior Research Methods, Instruments, \& Computers}}{36}{3}{402--407}.
\PrintBackRefs{\CurrentBib}

\bibitem [\protect \citeauthoryear {%
Papineni%
, Roukos%
, Ward%
\BCBL {}\ \BBA {} Zhu%
}{%
Papineni%
\ \protect \BOthers {.}}{%
{\protect \APACyear {2002}}%
}]{%
papineni-etal-2002-bleu}
\APACinsertmetastar {%
papineni-etal-2002-bleu}%
\begin{APACrefauthors}%
Papineni, K.%
, Roukos, S.%
, Ward, T.%
\BCBL {}\ \BBA {} Zhu, W\BHBI J.%
\end{APACrefauthors}%
\unskip\
\newblock
\APACrefYearMonthDay{2002}{}{}.
\newblock
{\BBOQ}\APACrefatitle {{BLEU}: {A} Method for Automatic Evaluation of Machine Translation} {{BLEU}: {A} method for automatic evaluation of machine translation}.{\BBCQ}
\newblock
\BIn{} \APACrefbtitle {{Proceedings of the Annual Meeting of the Association for Computational Linguistics}} {{Proceedings of the Annual Meeting of the Association for Computational Linguistics}}\ (\BPGS\ 311--318).
\PrintBackRefs{\CurrentBib}

\bibitem [\protect \citeauthoryear {%
Pennington%
, Socher%
\BCBL {}\ \BBA {} Manning%
}{%
Pennington%
\ \protect \BOthers {.}}{%
{\protect \APACyear {2014}}%
}]{%
pennington-etal-2014-glove}
\APACinsertmetastar {%
pennington-etal-2014-glove}%
\begin{APACrefauthors}%
Pennington, J.%
, Socher, R.%
\BCBL {}\ \BBA {} Manning, C.%
\end{APACrefauthors}%
\unskip\
\newblock
\APACrefYearMonthDay{2014}{}{}.
\newblock
{\BBOQ}\APACrefatitle {{G}lo{V}e: Global Vectors for Word Representation} {{G}lo{V}e: Global vectors for word representation}.{\BBCQ}
\newblock
\BIn{} \APACrefbtitle {{Proceedings of the Conference on Empirical Methods in Natural Language Processing}} {{Proceedings of the Conference on Empirical Methods in Natural Language Processing}}\ (\BPGS\ 1532--1543).
\PrintBackRefs{\CurrentBib}

\bibitem [\protect \citeauthoryear {%
Pirolli%
\ \BBA {} Card%
}{%
Pirolli%
\ \BBA {} Card%
}{%
{\protect \APACyear {1999}}%
}]{%
pirolli1999information}
\APACinsertmetastar {%
pirolli1999information}%
\begin{APACrefauthors}%
Pirolli, P.%
\BCBT {}\ \BBA {} Card, S.%
\end{APACrefauthors}%
\unskip\
\newblock
\APACrefYearMonthDay{1999}{}{}.
\newblock
{\BBOQ}\APACrefatitle {Information foraging} {Information foraging}.{\BBCQ}
\newblock
\APACjournalVolNumPages{Psychological review}{106}{4}{643}.
\PrintBackRefs{\CurrentBib}

\bibitem [\protect \citeauthoryear {%
Prud{'}hommeaux%
, van Santen%
\BCBL {}\ \BBA {} Gliner%
}{%
Prud{'}hommeaux%
\ \protect \BOthers {.}}{%
{\protect \APACyear {2017}}%
}]{%
prudhommeaux-etal-2017-vector}
\APACinsertmetastar {%
prudhommeaux-etal-2017-vector}%
\begin{APACrefauthors}%
Prud{'}hommeaux, E.%
, van Santen, J.%
\BCBL {}\ \BBA {} Gliner, D.%
\end{APACrefauthors}%
\unskip\
\newblock
\APACrefYearMonthDay{2017}{}{}.
\newblock
{\BBOQ}\APACrefatitle {Vector space models for evaluating semantic fluency in autism} {Vector space models for evaluating semantic fluency in autism}.{\BBCQ}
\newblock
\BIn{} \APACrefbtitle {{Proceedings of the Annual Meeting of the Association for Computational Linguistics}} {{Proceedings of the Annual Meeting of the Association for Computational Linguistics}}\ (\BPGS\ 32--37).
\PrintBackRefs{\CurrentBib}

\bibitem [\protect \citeauthoryear {%
Pyke%
, Pulliam%
\BCBL {}\ \BBA {} Charnov%
}{%
Pyke%
\ \protect \BOthers {.}}{%
{\protect \APACyear {1977}}%
}]{%
pyke1977optimal}
\APACinsertmetastar {%
pyke1977optimal}%
\begin{APACrefauthors}%
Pyke, G\BPBI H.%
, Pulliam, H\BPBI R.%
\BCBL {}\ \BBA {} Charnov, E\BPBI L.%
\end{APACrefauthors}%
\unskip\
\newblock
\APACrefYearMonthDay{1977}{}{}.
\newblock
{\BBOQ}\APACrefatitle {Optimal foraging: {A} selective review of theory and tests} {Optimal foraging: {A} selective review of theory and tests}.{\BBCQ}
\newblock
\APACjournalVolNumPages{{The Quarterly Review of Biology}}{52}{2}{137--154}.
\PrintBackRefs{\CurrentBib}

\bibitem [\protect \citeauthoryear {%
Sajjad%
\ \protect \BOthers {.}}{%
Sajjad%
\ \protect \BOthers {.}}{%
{\protect \APACyear {2022}}%
}]{%
sajjad-etal-2022-analyzing}
\APACinsertmetastar {%
sajjad-etal-2022-analyzing}%
\begin{APACrefauthors}%
Sajjad, H.%
, Durrani, N.%
, Dalvi, F.%
, Alam, F.%
, Khan, A.%
\BCBL {}\ \BBA {} Xu, J.%
\end{APACrefauthors}%
\unskip\
\newblock
\APACrefYearMonthDay{2022}{}{}.
\newblock
{\BBOQ}\APACrefatitle {Analyzing Encoded Concepts in Transformer Language Models} {Analyzing encoded concepts in transformer language models}.{\BBCQ}
\newblock
\BIn{} \APACrefbtitle {{Proceedings of the Conference of the North American Chapter of the Association for Computational Linguistics}} {{Proceedings of the Conference of the North American Chapter of the Association for Computational Linguistics}}\ (\BPGS\ 3082--3101).
\PrintBackRefs{\CurrentBib}

\bibitem [\protect \citeauthoryear {%
Suresh%
, Padua%
, Mukherjee%
\BCBL {}\ \BBA {} Rogers%
}{%
Suresh%
\ \protect \BOthers {.}}{%
{\protect \APACyear {2023}}%
}]{%
suresh2023behavioral}
\APACinsertmetastar {%
suresh2023behavioral}%
\begin{APACrefauthors}%
Suresh, S.%
, Padua, L.%
, Mukherjee, K.%
\BCBL {}\ \BBA {} Rogers, T\BPBI T.%
\end{APACrefauthors}%
\unskip\
\newblock
\APACrefYearMonthDay{2023}{}{}.
\newblock
{\BBOQ}\APACrefatitle {Behavioral estimates of conceptual structure are robust across tasks in humans but not large language models} {Behavioral estimates of conceptual structure are robust across tasks in humans but not large language models}.{\BBCQ}
\newblock
\APACjournalVolNumPages{arXiv preprint arXiv:2304.02754}{}{}{}.
\PrintBackRefs{\CurrentBib}

\bibitem [\protect \citeauthoryear {%
Touvron%
\ \protect \BOthers {.}}{%
Touvron%
\ \protect \BOthers {.}}{%
{\protect \APACyear {2023}}%
}]{%
touvron2023llama}
\APACinsertmetastar {%
touvron2023llama}%
\begin{APACrefauthors}%
Touvron, H.%
, Martin, L.%
, Stone, K.%
, Albert, P.%
, Almahairi, A.%
, ...%
\BCBL {}\ \BBA {} Scialom, T.%
\end{APACrefauthors}%
\unskip\
\newblock
\APACrefYearMonthDay{2023}{}{}.
\newblock
{\BBOQ}\APACrefatitle {{LLaMA} 2: Open foundation and fine-tuned chat models} {{LLaMA} 2: Open foundation and fine-tuned chat models}.{\BBCQ}
\newblock
\APACjournalVolNumPages{arXiv preprint arXiv:2307.09288}{}{}{}.
\PrintBackRefs{\CurrentBib}

\bibitem [\protect \citeauthoryear {%
Troyer%
, Moscovitch%
\BCBL {}\ \BBA {} Winocur%
}{%
Troyer%
\ \protect \BOthers {.}}{%
{\protect \APACyear {1997}}%
}]{%
troyer1997clustering}
\APACinsertmetastar {%
troyer1997clustering}%
\begin{APACrefauthors}%
Troyer, A\BPBI K.%
, Moscovitch, M.%
\BCBL {}\ \BBA {} Winocur, G.%
\end{APACrefauthors}%
\unskip\
\newblock
\APACrefYearMonthDay{1997}{}{}.
\newblock
{\BBOQ}\APACrefatitle {Clustering and switching as two components of verbal fluency: evidence from younger and older healthy adults.} {Clustering and switching as two components of verbal fluency: evidence from younger and older healthy adults.}{\BBCQ}
\newblock
\APACjournalVolNumPages{Neuropsychology}{11}{1}{138}.
\PrintBackRefs{\CurrentBib}

\bibitem [\protect \citeauthoryear {%
Troyer%
, Moscovitch%
, Winocur%
, Alexander%
\BCBL {}\ \BBA {} Stuss%
}{%
Troyer%
\ \protect \BOthers {.}}{%
{\protect \APACyear {1998}}%
}]{%
troyer1998clustering}
\APACinsertmetastar {%
troyer1998clustering}%
\begin{APACrefauthors}%
Troyer, A\BPBI K.%
, Moscovitch, M.%
, Winocur, G.%
, Alexander, M\BPBI P.%
\BCBL {}\ \BBA {} Stuss, D.%
\end{APACrefauthors}%
\unskip\
\newblock
\APACrefYearMonthDay{1998}{}{}.
\newblock
{\BBOQ}\APACrefatitle {Clustering and switching on verbal fluency: The effects of focal frontal-and temporal-lobe lesions} {Clustering and switching on verbal fluency: The effects of focal frontal-and temporal-lobe lesions}.{\BBCQ}
\newblock
\APACjournalVolNumPages{Neuropsychologia}{36}{6}{499--504}.
\PrintBackRefs{\CurrentBib}

\bibitem [\protect \citeauthoryear {%
Wilcox%
, Gauthier%
, Hu%
, Qian%
\BCBL {}\ \BBA {} Levy%
}{%
Wilcox%
\ \protect \BOthers {.}}{%
{\protect \APACyear {2020}}%
}]{%
wilcox2020predictive}
\APACinsertmetastar {%
wilcox2020predictive}%
\begin{APACrefauthors}%
Wilcox, E\BPBI G.%
, Gauthier, J.%
, Hu, J.%
, Qian, P.%
\BCBL {}\ \BBA {} Levy, R.%
\end{APACrefauthors}%
\unskip\
\newblock
\APACrefYearMonthDay{2020}{}{}.
\newblock
{\BBOQ}\APACrefatitle {On the predictive power of neural language models for human real-time comprehension behavior} {On the predictive power of neural language models for human real-time comprehension behavior}.{\BBCQ}
\newblock
\APACjournalVolNumPages{arXiv preprint arXiv:2006.01912}{}{}{}.
\PrintBackRefs{\CurrentBib}

\bibitem [\protect \citeauthoryear {%
Zemla%
\ \BBA {} Austerweil%
}{%
Zemla%
\ \BBA {} Austerweil%
}{%
{\protect \APACyear {2018}}%
}]{%
zemla2018estimating}
\APACinsertmetastar {%
zemla2018estimating}%
\begin{APACrefauthors}%
Zemla, J\BPBI C.%
\BCBT {}\ \BBA {} Austerweil, J\BPBI L.%
\end{APACrefauthors}%
\unskip\
\newblock
\APACrefYearMonthDay{2018}{}{}.
\newblock
{\BBOQ}\APACrefatitle {Estimating semantic networks of groups and individuals from fluency data} {Estimating semantic networks of groups and individuals from fluency data}.{\BBCQ}
\newblock
\APACjournalVolNumPages{{Computational Brain \& Behavior}}{1}{}{36--58}.
\PrintBackRefs{\CurrentBib}

\end{thebibliography}

\end{document}